\documentclass{article}
\usepackage{spconf,amsmath,graphicx}
\usepackage{xcolor}	
\usepackage{array,multirow}
\usepackage{booktabs}
\usepackage{comment}
\usepackage{url}
\usepackage{pifont}

\title{CLICKER: Attention-Based Cross-Lingual Commonsense Knowledge Transfer}
\name{Ruolin Su$^{1*}$, Zhongkai Sun$^2$, Sixing Lu$^2$, Chengyuan Ma$^2$, Chenlei Guo$^2$\thanks{$^*$~Work done during internship at Amazon Alexa AI.}}
\address{$^1$Georgia Institute of Technology \ $^2$Amazon Alexa AI\\
ruolinsu@gatech.edu, \{zhongkas, cynthilu, mchengyu, guochenl\}@amazon.com}
\begin{document}
%
\maketitle
\begin{abstract}

Recent advances in cross-lingual commonsense reasoning (CSR) are facilitated by the development of multilingual pre-trained models (mPTMs).
While mPTMs show the potential to encode commonsense knowledge for different languages, transferring commonsense knowledge learned in large-scale English corpus to other languages is challenging. 
To address this problem, we propose the attention-based Cross-LIngual Commonsense Knowledge transfER (CLICKER) framework, which minimizes the performance gaps between English and non-English languages in commonsense question-answering tasks. 
CLICKER effectively improves commonsense reasoning for non-English languages by differentiating non-commonsense knowledge from commonsense knowledge.
Experimental results on public benchmarks demonstrate that CLICKER achieves remarkable improvements in the cross-lingual CSR task for languages other than English.
\end{abstract}
\begin{keywords}
Commonsense reasoning, multilingual, pre-trained language model, knowledge extraction, self-attention
\end{keywords}

\begin{figure*}[!htb]
     \centering
     \includegraphics[width=\textwidth,height=6.8cm]{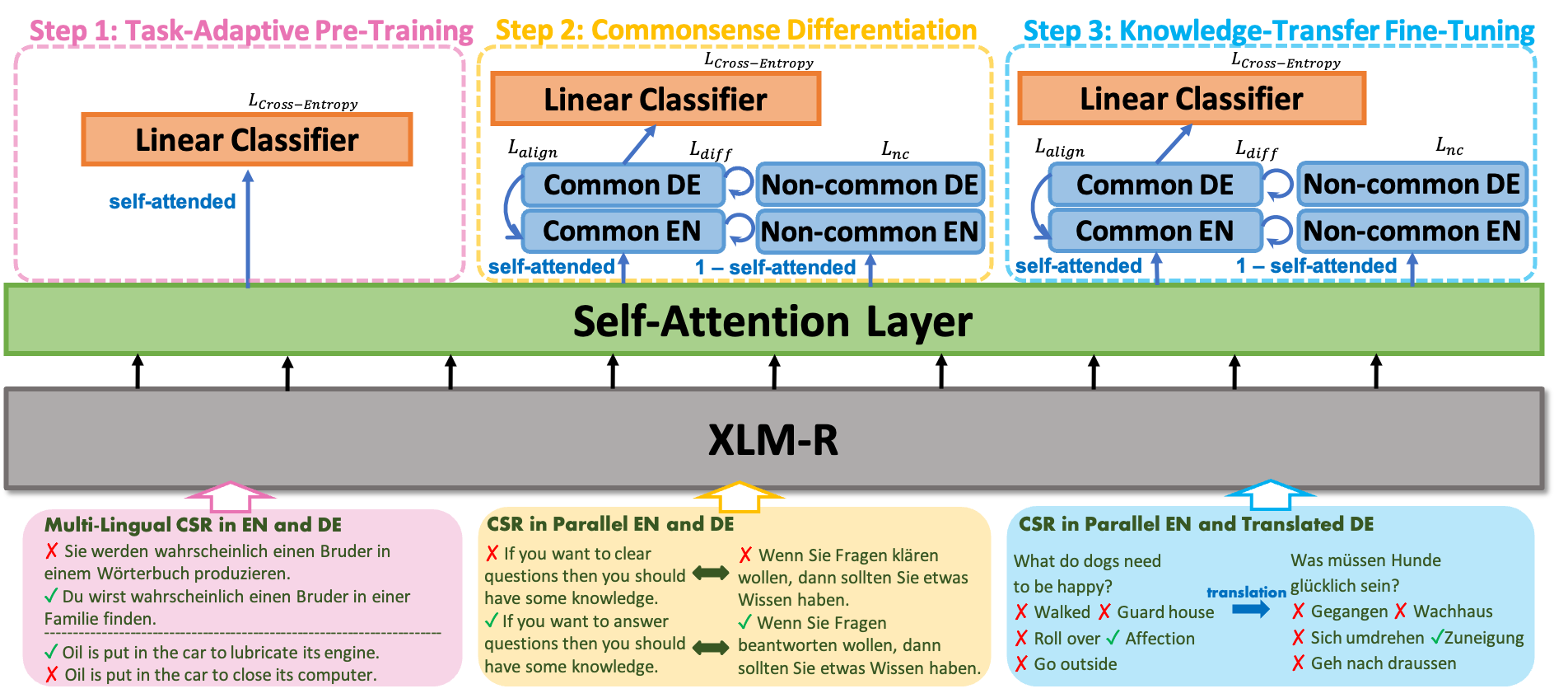} %
     \caption{The overview of CLICKER framework, which is trained in three steps with joint objectives for cross-lingual commonsense reasoning. Parameters in XLM-R and self-attention layers are shared in all steps. Each \ding{52} or \ding{56} represents whether the following choice is commonsense reasonable or not, respectively.}\label{Fig:model}
\end{figure*}

\section{Introduction}
 Commonsense reasoning (CSR) relies on shared and unchanged knowledge across different languages and cultures to help computers understand and interact with humans naturally~\cite{davis2015commonsense}. CSR is a crucial problem in natural language processing that has been proved important for artificial intelligence (AI) systems~\cite{talmor-etal-2019-commonsenseqa, storks2019recent}.

Cross-lingual CSR aims to reason commonsense across languages,
 which is the key to bridging the language barrier in natural language understanding and generalizing CSR to a broader scope~\cite{hu-2020-xtreme}. Recently, several cross-lingual datasets are proposed amidst the surging interests in cross-lingual CSR, \textit{e.g.} XCOPA~\cite{ponti-2020-xcopa}, X-CSQA~\cite{lin-etal-2021-xcsr}, and X-CODAH~\cite{lin-etal-2021-xcsr}.
Multilingual pre-trained language models (mPTMs) based on the Transformer~\cite{NIPS2017_3f5ee243}, such as mBERT~\cite{DBLP:journals/corr/abs-1810-04805}, XLM~\cite{lample2019cross}, XLM-R~\cite{conneau2019unsupervised} and InfoXLM~\cite{chi2020infoxlm}, have also been demonstrated to have
the potentials of conducting CSR in multiple languages~\cite{conneau2018xnli,hu-2020-xtreme,ponti-2020-xcopa,lin-etal-2021-xcsr}. 
The performance of mPTMs for non-English CSR, however, is typically worse than that for English CSR due to the lack of non-English data for training~\cite{yeo2018machine,conneau2019unsupervised,ponti-2020-xcopa}.
 Furthermore, mPTMs have raised concerns about their ability to transfer commonsense knowledge across languages, as they do not 1) differentiate  between commonsense and non-commonsense knowledge and 2) improve the CSR for any specific language in multilingual scenarios~\cite{conneau2019unsupervised}.

To address the above issues, we propose a Cross-LIngual Commonsense Knowledge transfER (\textbf{CLICKER}) to bridge the performance gap of using mPTMs for CSR between the source (\textbf{English}) and the target (\textbf{non-English}) language by eliciting commonsense knowledge explicitly via cross-lingual task-adaptive pre-training~\cite{gururangan2020don}.
 Specifically, CLICKER is a three-step framework based on XLM-R~\cite{conneau2019unsupervised}. 
 First, we conduct task-adaptive pre-training on the multilingual commonsense corpora to enable XLM-R to perform the CSR task better. 
 In this process, the self-attention~\cite{NIPS2017_3f5ee243} mechanism is adopted to obtain multilingual embeddings for CSR. 
 Second, we distinguish between commonsense and non-commonsense knowledge by jointly optimizing their similarities with bilingual and parallel data. 
 Third, the extracted commonsense knowledge representation is further fine-tuned on the downstream cross-lingual CSR tasks.
 
 Experimental results demonstrate that our approach significantly reduces the performance discrepancies between English and German on CSR.
 Moreover, it outperforms XLM-R baselines on both X-CSQA and X-CODAH benchmarks~\cite{lin-etal-2021-xcsr}.
 Further analysis indicates that CLICKER can extract cross-lingual commonsense representations more effectively, and with better interpretability.

\section{Method}
This section introduces the CLICKER model based on XLM-R~\cite{conneau2019unsupervised} for cross-lingual CSR.
As illustrated in Figure~\ref{Fig:model},
CLICKER extracts commonsense knowledge from English CSR to help non-English CSR\footnote{In this paper, we take German as an example of a foreign language that is not up to par with English for CSR.} in three steps: 1) task-adaptive pre-training, 2) commonsense differentiation, and 3) knowledge-transfer fine-tuning.

\subsection{Problem Definition}

The CSR task aims to select one from multiple choices that are most reasonable in commonsense given the previous statement or question.
 For example, the plausible choice of answer to \textit{``What is a great place to lay in the sun?''}  is \textit{``beach''} rather than \textit{``in the basement''} or \textit{``solar system''}.
 Denoting a set of choices for CSR as $\mathbf{S}^{(j)}$'s, $j \in$ [1,$\dots$, $|C|$], where the number of choices for each input as $|C|$, the goal is to predict the common sense choice:
 \begin{equation}
    \begin{aligned}\label{eq:object}
\tilde{y}  = \text{argmax}_j P(y=j | \{\mathbf{S}^{(1)}, \dots, \mathbf{S}^{(|C|)}\})\\
\end{aligned}
\end{equation}

\subsection{Step One: Task-Adaptive Pre-Training}\label{sec:step1}
Task-adaptive pre-training uses self-attention to enable the XLM-R to learn representations for the CSR task.
Specifically, the input is a tokenized utterance, \textit{i.e.} $\mathbf{S}_i=\{\text{[CLS]},q_i^{(1)},$
$\dots,q_i^{(k)},\text{[SEP]}\}$, where $i\in [1,\dots,N]$, $N$ is the size of the dataset, and $k$ is the length of sequence.
A self-attention layer is built on top of the Transformer to obtain attentions toward commonsense knowledge from the Transformer's pooled output states. 
The self-attended outputs are optimized by the Cross-Entropy loss through a multiple-choice classifier to select commonsense-reasonable choices.
Our model is trained on multilingual CSR datasets containing examples of both English (\textbf{EN}) and German (\textbf{DE}).

\subsection{Step Two: Commonsense Differentiation}\label{sec:step2}
In this step, the representation of commonsense knowledge shared across languages is differentiated from the non-commonsense representation using EN and DE datasets in parallel.
The inputs are similar to those in Sec~\ref{sec:step1}, while inputs with the same semantics in different languages are mapped together. 
We note here that the parallel datasets are not necessarily restricted to CSR datasets, but can be generalized to any bilingual datasets for mapping semantics of English and the non-English language, e.g. bilingual dictionaries or textbooks.

The output states of the Transformer are pooled and weighted by the self-attention layer followed by a linear projection, being extracted as \textbf{commonsense} embeddings $X_i$ and \textbf{non-commonsense} embeddings $\tilde{X_i}$, respectively.

\begin{equation}
    \begin{aligned}\label{eq:residual}
X_i = & FFN(\text{Attention}(\mathbf{O}_i))
\end{aligned}
\end{equation}
\begin{equation}
    \begin{aligned}\label{eq:attn}
\tilde{X_i} = & FFN(1-\text{Attention}(\mathbf{O}_i))
\end{aligned}
\end{equation}
where $\mathbf{O}_i$ are output hidden states from the last layer of the Transformer for the $i$-\textit{th} input, and $FFN$ represents a \textit{Feed-Forward} layer.
For brevity, we omit index $i$ in the following equations.

We use $X^{EN}$ and $X^{DE}$ to denote commonsense embeddings of English and German inputs
, respectively.
Similarly, $\tilde{X}^{EN}$ and $\tilde{X}^{DE}$ represent non-commonsense embeddings.
Knowledge mapping is made by measuring the similarities between commonsense embeddings and non-commonsense embeddings.
Specifically, we maximize the cosine similarity between English and German embeddings that share the same and valid commonsense knowledge, \textit{i.e.} $X^{EN_{j^*}}$ and $X^{DE_{j^*}}$, as in Eq.~(\ref{eq:common1}). And we minimize the cosine similarity between
$X^{EN_{j^*}}$ and $X^{EN_j}$, as in Eq.~(\ref{eq:common2}).
$j^*$ is the index of the choice that is reasonable in commonsense, $j \in$ [1,$\dots$, $|C|$] and $j \neq {j^*}$.
Such that similar commonsense knowledge in both languages is projected into the same position in the semantic representation space.
\begin{equation}
\label{eq:common1}
\mathcal{L}_{align} =  1 - \cos(X^{EN_{j^*}}, X^{DE_{j^*}})
\end{equation}
\begin{equation}
\begin{aligned}\label{eq:common2}
\mathcal{L}_{diff} &= \sum_{j= 1, j \neq {j^*}}^{|C|} (\text{max}(0, \cos(X^{EN_{j^*}}, X^{EN_j})) \\
 &+~\text{max}(0,\cos(X^{DE_{j^*}}, X^{DE_j})) )\\
\end{aligned}
\end{equation}

On the other hand, the non-commonsense embeddings represent knowledge unrelated to cross-lingual commonsense. 
Assuming the correct choice and other incorrect choices associated with the same question share similar non-commonsense knowledge, we maximize the intra-language cosine similarity of non-commonsense embeddings. 
Moreover, the correct choice of different languages should share the same non-commonsense knowledge so that we maximize inter-language cosine similarity jointly, as defined in Eq.~(\ref{eq:res}).
\begin{equation}
\begin{aligned}\label{eq:res}
\mathcal{L}_{nc}  &= \sum_{j= 1, j \neq {j^*}}^{|C|} ( 1 - \cos(\tilde{X_i}^{EN_{j^*}}, \tilde{X_i}^{EN_j}) ) \\
& + \sum_{j= 1, j \neq {j^*}}^{|C|} ( 1 - \cos(\tilde{X_i}^{DE_{j^*}}, \tilde{X_i}^{DE_j}))\\
& + 1 - \cos(\tilde{X_i}^{EN_{j^*}}, \tilde{X_i}^{DE_{j^*}}) \\
\end{aligned}
\end{equation}
All the losses above and the Cross-Entropy loss are optimized
as the joint training objective of the cross-lingual CSR.
We use output commonsense embeddings $X^{DE_{j^*}}$ and $X^{DE_{j}}$ to calculate the Cross-Entropy loss.

\subsection{Step Three: Knowledge-Transfer Fine-Tuning}\label{sec:step3}
Finally, our model is fine-tuned by the training objectives similar to Sec~\ref{sec:step2} for evaluating CSR on the multiple-choice question-answering (QA) and the clause-selection tasks, leveraging parallel CSR datasets of English (\textbf{EN}) and German translated from English (\textbf{EN\_DE}) as inputs.
Different from previous steps, each input of XLM-R is the concatenation of a question and a choice of answer which are then split into tokens with additional special ones, \textit{i.e.} $\mathbf{S}_i=\{\text{[CLS]},q_i^{(1)},\dots,q_i^{(m)},\text{[SEP]},\text{[CLS\_Q]},a_i^{(1)},\dots,$ $a_i^{(n)},\text{[SEP]}\}$, where [CLS\_Q] is the beginning special token of the answer spans, $q_i$ and $a_i$ are tokens of the question and answer, and $m$, $n$ are numbers of question and answer tokens, respectively. 

\section{Experiments and Analyses}
We use English and German subsets of Mickey Corpus~\cite{lin-etal-2021-xcsr} for Step 1 to warm up the multilingual language model for cross-lingual CSR tasks.
Then we take advantage of parallel corpora of English and German in the Mickey Corpus again for Step 2 to obtain their semantic mappings and differentiate commonsense and non-commonsense embeddings.
For Step 3, the CLICKER model is fine-tuned on the English and machine-translated German training set of X-CSQA and X-CODAH~\cite{lin-etal-2021-xcsr}, which are in the style of multiple-choice QA and selection of appropriate clauses, respectively. 

We compare our model with the multilingual contrastive pre-training (MCP)~\cite{lin-etal-2021-xcsr} model based on XLM-R$_B$~\cite{conneau2019unsupervised}.
MCP model is trained on permuted Mickey Corpus for multilingual contrastive training and fine-tunes on cross-lingual CSR training set in English only.
Instead, we re-implement it to train on the combination of English and German Mickey Corpus. Then we fine-tune it on both English and machine-translated German CSR training sets and evaluate it on the test set in German to make a fair comparison with our method.

The following subsections describe the experimental results and analyze CLICKER models on the cross-lingual CSR benchmarks X-CSQA and X-CODAH in German.
Note that our experiments are conducted for commonsense knowledge transfer from English to German, but the approach can be extended to other languages.

\subsection{Experimental Results}

Table~\ref{xcsqa} shows the test accuracy of baselines and CLICKER models for CSR in German. 
Different combinations of losses are applied in experiments for optimizing commonsense differentiation. 
We observe consistent improvements with our three-step framework by extracting commonsense knowledge with self-attentions (\textit{i.e.} CLICKER - \textit{base}) on both datasets compared to baselines.


Results show that the \textit{align} loss further improves the base CLICK model on X-CSQA.
And the \textit{non-commonsense (nc)} loss is proved effective on both datasets.
The best performance on X-CSQA is achieved when using the \textit{align} loss with or without the \textit{diff} loss, which shows that lining up embeddings in English and German with the same commonsense knowledge dominates the performance of CSR. 
Besides, the model with \textit{align} and \textit{nc} loss is slightly inferior to the model with \textit{nc} loss only on X-CSQA. 
 On X-CODAH, our CLICK models perform the best with the \textit{nc} loss which maximizes the cosine similarity of non-commonsense embeddings, improving 1.6\% on accuracy.

\begin{table}
\centering
\begin{tabular}{{p{0.7\linewidth}p{0.22\linewidth}}}
\hline
\textbf{Models} & \textbf{Acc} \\
\hline
\multicolumn{2}{ l }{\textit{X-CSQA}}\\
\hline
MCP(XLM-R$_B$)*~\cite{lin-etal-2021-xcsr} & 48.8 \\

CLICKER - \textit{base} & 49.6 (+0.8) \\

CLICKER - \textit{align} & \textbf{50.6 (+1.8)} \\

CLICKER - \textit{align+diff} & \textbf{50.6 (+1.8)} \\

CLICKER - \textit{nc} & 49.8 (+1.0) \\

CLICKER - \textit{align+nc} & 49.6 (+0.8) \\


\hline
\multicolumn{2}{ l }{\textit{X-CODAH}}\\
\hline
MCP(XLM-R$_B$)*~\cite{lin-etal-2021-xcsr} & 49.2 \\

CLICKER - \textit{base} & 50.2 (+1.0) \\

CLICKER - \textit{align} & 49.6 (+0.4) \\
CLICKER - \textit{align+diff} &  50.3 (+1.1)\\

CLICKER - \textit{nc} & \textbf{50.8 (+1.6)} \\

CLICKER - \textit{align+nc} &  49.6 (+0.4)\\



\hline
\end{tabular}
\caption{\label{xcsqa}
Accuracy on the test set of X-CSQA and X-CODAH in German. MCP(XLM-R$_B$)* model is trained in English and machine-translated German. The $align$, $diff$, $nc$ refer to the objectives in equation (\ref{eq:common1}), (\ref{eq:common2}), and (\ref{eq:res}), respectively.
}
\end{table}


\subsection{Discussion} 
Our models address the alignment of extracted embeddings with various combinations of objectives.
The fact that \textit{align+nc} loss is not as good as \textit{nc} loss alone suggests a conflict between aligning the commonsense embeddings and aligning the non-commonsense embeddings.
This can be explained as both objectives aiming to maximize the cosine similarity of embeddings,  making it harder for the model to discern different commonsense knowledge in them. 
From the best accuracy achieved on two datasets, we conjecture the quality of commonsense embeddings (optimized by \textit{align} and \textit{diff} losses) dominates CSR on X-CSQA, while non-commonsense embeddings (optimized by \textit{nc} loss) dominates that on X-CODAH.
The reason for this may be extracting commonsense knowledge for clause selection in X-CODAH is more challenging than multiple-choice QA in X-CSQA, whereas separating the non-commonsense embeddings help the multiple-choice classifier understand the commonsense portion with less noise.
We also observe that using \textit{align} and \textit{nc} losses together is not the best practice according to our experiments. 
It suggests that jointly optimizing both objectives makes it more difficult for the multiple-choice classifier to predict correctly, as correct choices are pushed closer to incorrect ones.

\textbf{Commonsense v.s. Non-commonsense.} To investigate the effectiveness of our learned commonsense embeddings, we evaluate the accuracy of our CLICKER models on the X-CSQA dev set predicted by commonsense embeddings or non-commonsense embeddings. As seen in Table~\ref{embed}, the performance of commonsense embeddings is significantly better than that of non-commonsense embeddings. 
It is as expected, as our models are trained with cross-lingual CSR objectives to discern commonsense embeddings, while maximizing the similarity of non-commonsense embeddings. 
Non-commonsense embeddings can induce confusion for CSR, such that combining both embeddings performs worse than using commonsense embeddings only.
\begin{figure}[!htb]
    \centering
     \includegraphics[width=.86\linewidth]{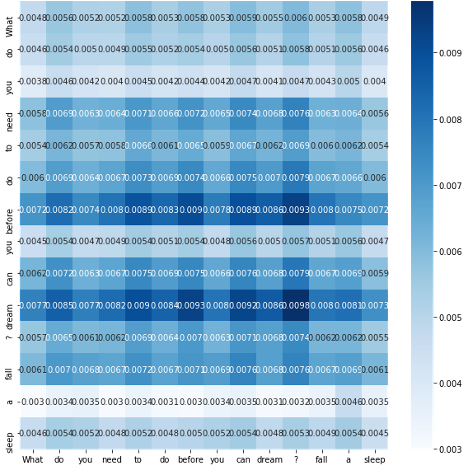}
     \caption{Attention head of the self-attention layer. The given example is from X-CSQA.}\label{Fig:Data2}
\end{figure}

\textbf{Does self-attention imply commonsense knowledge?} We assume that self-attentions in our models can appropriately attend to tokens that affect the plausibility of commonsense. 
Figure~\ref{Fig:Data2} is the heatmap of the attention head in the self-attention layer evaluated on an example ``\textit{What do you need to be before you can dream?}'' from X-CSQA. 
It's noteworthy to see that  attention weights are given more to commonsense-related tokens, such as 
``\textit{before}'', ``\textit{dream}'' and ``\textit{sleep}'' tokens.  
A similar phenomenon is observed on X-CODAH as well.
These tokens are weighted to generate commonsense embeddings and help our model improve accuracy and interpretability of reasoning commonsense knowledge.
\begin{table}[!t]
\centering
\begin{tabular}{ll}
\hline
\textbf{Classifier Input} & \textbf{Dev Acc} \\
\hline
Commonsense & \textbf{47.8} \\
Non-Commonsense & 11.0 \\
Commonsense + Non-Commonsense & 47.6 \\
\hline
\end{tabular}
\caption{\label{embed}
Dev accuracy on X-CSQA taking commonsense or non-commonsense embeddings as inputs for the classifier.
}
\end{table}
\section{Conclusion}
In this paper, we propose a cross-lingual framework CLICKER for commonsense reasoning.
Experiments on X-CSQA and X-CODAH demonstrate the effectiveness of CLICKER in cross-lingual commonsense reasoning as it not only reduces performance discrepancies of commonsense reasoning between English and non-English languages but also improves the interpretability of commonsense knowledge across languages.
The potential of our approach to be generalized to other low-resource languages will be beneficial for alleviating data scarcity in cross-lingual commonsense reasoning.

\vfill\pagebreak

\bibliographystyle{IEEEbib}
\bibliography{refs, strings}

\begin{thebibliography}{10}

\bibitem{davis2015commonsense}
Ernest Davis and Gary Marcus,
\newblock ``Commonsense reasoning and commonsense knowledge in artificial
  intelligence,''
\newblock {\em Communications of the ACM}, vol. 58, no. 9, pp. 92--103, 2015.

\bibitem{talmor-etal-2019-commonsenseqa}
Alon Talmor, Jonathan Herzig, Nicholas Lourie, and Jonathan Berant,
\newblock ``{C}ommonsense{QA}: A question answering challenge targeting
  commonsense knowledge,''
\newblock in {\em Proceedings of the 2019 Conference of the North {A}merican
  Chapter of the Association for Computational Linguistics: Human Language
  Technologies, Volume 1 (Long and Short Papers)}, Minneapolis, Minnesota, June
  2019, pp. 4149--4158, Association for Computational Linguistics.

\bibitem{storks2019recent}
Shane Storks, Qiaozi Gao, and Joyce~Y Chai,
\newblock ``Recent advances in natural language inference: A survey of
  benchmarks, resources, and approaches,''
\newblock {\em arXiv preprint arXiv:1904.01172}, 2019.

\bibitem{hu-2020-xtreme}
Junjie Hu, Sebastian Ruder, Aditya Siddhant, Graham Neubig, Orhan Firat, and
  Melvin Johnson,
\newblock ``{XTREME:} {A} massively multilingual multi-task benchmark for
  evaluating cross-lingual generalization,''
\newblock {\em CoRR}, vol. abs/2003.11080, 2020.

\bibitem{ponti-2020-xcopa}
Edoardo~Maria Ponti, Goran Glavas, Olga Majewska, Qianchu Liu, Ivan Vulic, and
  Anna Korhonen,
\newblock ``{XCOPA:} {A} multilingual dataset for causal commonsense
  reasoning,''
\newblock {\em CoRR}, vol. abs/2005.00333, 2020.

\bibitem{lin-etal-2021-xcsr}
Bill~Yuchen Lin, Seyeon Lee, Xiaoyang Qiao, and Xiang Ren,
\newblock ``Common sense beyond english: Evaluating and improving multilingual
  language models for commonsense reasoning,''
\newblock in {\em Proceedings of the 59th Annual Meeting of the Association for
  Computational Linguistics (ACL-IJCNLP 2021)}, 2021.

\bibitem{NIPS2017_3f5ee243}
Ashish Vaswani, Noam Shazeer, Niki Parmar, Jakob Uszkoreit, Llion Jones,
  Aidan~N Gomez, \L~ukasz Kaiser, and Illia Polosukhin,
\newblock ``Attention is all you need,''
\newblock in {\em Advances in Neural Information Processing Systems}, I.~Guyon,
  U.~V. Luxburg, S.~Bengio, H.~Wallach, R.~Fergus, S.~Vishwanathan, and
  R.~Garnett, Eds. 2017, vol.~30, Curran Associates, Inc.

\bibitem{DBLP:journals/corr/abs-1810-04805}
Jacob Devlin, Ming{-}Wei Chang, Kenton Lee, and Kristina Toutanova,
\newblock ``{BERT:} pre-training of deep bidirectional transformers for
  language understanding,''
\newblock {\em CoRR}, vol. abs/1810.04805, 2018.

\bibitem{lample2019cross}
Guillaume Lample and Alexis Conneau,
\newblock ``Cross-lingual language model pretraining,''
\newblock {\em arXiv preprint arXiv:1901.07291}, 2019.

\bibitem{conneau2019unsupervised}
Alexis Conneau, Kartikay Khandelwal, Naman Goyal, Vishrav Chaudhary, Guillaume
  Wenzek, Francisco Guzm{\'{a}}n, Edouard Grave, Myle Ott, Luke Zettlemoyer,
  and Veselin Stoyanov,
\newblock ``Unsupervised cross-lingual representation learning at scale,''
\newblock {\em CoRR}, vol. abs/1911.02116, 2019.

\bibitem{chi2020infoxlm}
Zewen Chi, Li~Dong, Furu Wei, Nan Yang, Saksham Singhal, Wenhui Wang, Xia Song,
  Xian-Ling Mao, Heyan Huang, and Ming Zhou,
\newblock ``{I}nfo{XLM}: An information-theoretic framework for cross-lingual
  language model pre-training,''
\newblock in {\em Proceedings of the 2021 Conference of the North American
  Chapter of the Association for Computational Linguistics: Human Language
  Technologies}, Online, June 2021, pp. 3576--3588, Association for
  Computational Linguistics.

\bibitem{conneau2018xnli}
Alexis Conneau, Ruty Rinott, Guillaume Lample, Adina Williams, Samuel~R.
  Bowman, Holger Schwenk, and Veselin Stoyanov,
\newblock ``Xnli: Evaluating cross-lingual sentence representations,''
\newblock in {\em Proceedings of the 2018 Conference on Empirical Methods in
  Natural Language Processing}. 2018, Association for Computational
  Linguistics.

\bibitem{yeo2018machine}
Jinyoung Yeo, Geungyu Wang, Hyunsouk Cho, Seungtaek Choi, and Seung-won Hwang,
\newblock ``Machine-translated knowledge transfer for commonsense causal
  reasoning,''
\newblock in {\em Proceedings of the AAAI Conference on Artificial
  Intelligence}, 2018, vol.~32.

\bibitem{gururangan2020don}
Suchin Gururangan, Ana Marasovi{\'c}, Swabha Swayamdipta, Kyle Lo, Iz~Beltagy,
  Doug Downey, and Noah~A Smith,
\newblock ``Don't stop pretraining: adapt language models to domains and
  tasks,''
\newblock {\em arXiv preprint arXiv:2004.10964}, 2020.

\end{thebibliography}

\end{document}